\documentclass[10pt,twocolumn,letterpaper]{article}

\usepackage{cvpr}              

\usepackage{graphicx}
\usepackage{amsmath}
\usepackage{amssymb}
\usepackage{stmaryrd}
\usepackage{booktabs}
\usepackage{multirow}

\usepackage{enumitem}
\setenumerate[1]{itemsep=0pt,partopsep=0pt,parsep=\parskip,topsep=5pt}
\setitemize[1]{itemsep=0pt,partopsep=0pt,parsep=\parskip,topsep=5pt}
\setdescription{itemsep=0pt,partopsep=0pt,parsep=\parskip,topsep=5pt}

\usepackage[labelformat=simple]{subcaption}

\usepackage[pagebackref,breaklinks,colorlinks]{hyperref}

\usepackage{marvosym}

\usepackage[capitalize]{cleveref}
\crefname{section}{Sec.}{Secs.}
\Crefname{section}{Section}{Sections}
\Crefname{table}{Table}{Tables}
\crefname{table}{Tab.}{Tabs.}

\def\cvprPaperID{4040} 
\def\confName{CVPR}
\def\confYear{2023}

\newcommand*{\affaddr}[1]{#1} 
\newcommand*{\affmark}[1][*]{\textsuperscript{#1}}

\makeatletter
\newcommand{\printfnsymbol}[1]{\textsuperscript{\@fnsymbol{#1}}}
\makeatother

\makeatletter
\def\thanks#1{\protected@xdef\@thanks{\@thanks
        \protect\footnotetext{#1}}}
\makeatother
\begin{document}

\title{Understanding the Robustness of 3D Object Detection with Bird's-Eye-View Representations in Autonomous Driving}

\author{
Zijian Zhu\affmark[1]\textsuperscript{*}\quad 
Yichi Zhang\affmark[2,5]\textsuperscript{*}\quad 
Hai Chen\affmark[3]\quad 
Yinpeng Dong\affmark[2]\textsuperscript{\Letter}\quad \\
Shu Zhao\affmark[3]\quad
Wenbo Ding\affmark[4]\quad
Jiachen Zhong\affmark[4]\quad
Shibao Zheng\affmark[1]\textsuperscript{\Letter}\thanks{\textsuperscript{*} Equal Contribution. \textsuperscript{\Letter}~Corresponding authors. This work was done when Zijian Zhu and Hai Chen were visiting Tsinghua University.}\quad 
\\\affaddr{\affmark[1] 
Institute of Image Communication and Network Engineering, Shanghai Jiao Tong University} 
\\\affaddr{\affmark[2] 
Dept. of Comp. Sci. and Tech., Institute for AI, THBI Lab, BNRist Center, Tsinghua University} 
\\\affaddr{\affmark[3]
Key Laboratory of Intelligent Computing and Signal Processing, Ministry of Education, \\
School of Computer Science and Technology, Anhui University, \\
Information Materials and Intelligent Sensing Laboratory of Anhui Province}
\\\affaddr{\affmark[4] SAIC Motor AI Lab} \;
\affaddr{\affmark[5] Zhongguancun Laboratory}
\\
\hspace{-1.5ex}\tt\small{\{zzj403,sbzh\}@sjtu.edu.cn, zyc22@mails.tsinghua.edu.cn, dongyinpeng@tsinghua.edu.cn}\\
\tt\small{\{chber$\_$ahu,zhaoshuzs2002\}@hotmail.com, \{dingwenbo,zhongjiachen\}@saicmotor.com}\\
}

\maketitle

\begin{abstract}
3D object detection is an essential perception task in autonomous driving to understand the environments. The Bird's-Eye-View (BEV) representations have significantly improved the performance of 3D detectors with camera inputs on popular benchmarks. However, there still lacks a systematic understanding of the robustness of these vision-dependent BEV models, which is closely related to the safety of autonomous driving systems. In this paper, we evaluate the natural and adversarial robustness of various representative models under extensive settings, to fully understand their behaviors influenced by explicit BEV features compared with those without BEV. In addition to the classic settings, we propose a 3D consistent patch attack by applying adversarial patches in the 3D space to guarantee the spatiotemporal consistency, which is more realistic for the scenario of autonomous driving. With substantial experiments, we draw several findings: 1) BEV models tend to be more stable than previous methods under different natural conditions and common corruptions due to the expressive spatial representations; 2) BEV models are more vulnerable to adversarial noises, mainly caused by the redundant BEV features; 3) Camera-LiDAR fusion models have superior performance under different settings with multi-modal inputs, but BEV fusion model is still vulnerable to adversarial noises of both point cloud and image. These findings alert the safety issue in the applications of BEV detectors and could facilitate the development of more robust models. Code available at \url{https://github.com/zzj403/BEV_Robust}.
\end{abstract}

\vspace{-4ex}
\section{Introduction}

Autonomous driving systems have great demand for reliable 3D object detection models~\cite{grigorescu2020survey}, which aim to predict 3D bounding boxes and categories of road objects, in order to understand the surroundings. To extract holistic representations in the 3D space, the Bird's-Eye-View (BEV) is commonly adopted as a unified representation~\cite{li2022delving}, since it contains both locations and semantic features of objects without being affected by occlusion, and shows promise for various 3D perception tasks in autonomous driving, such as map restoration~\cite{philion2020lift,ng2020bev}. Although being broadly used for LiDAR point clouds~\cite{zhou2018voxelnet,lang2019pointpillars}, the BEV representation has recently achieved great success for 3D object detection with multiple cameras, arousing tremendous attention from both industry and academia due to low cost of camera sensors and better exploitation of semantic information in images.
These \emph{vision-dependent BEV models}\footnote{In this paper, we use the term vision-dependent BEV models to indicate both camera-only and LiDAR-camera fusion BEV models.} typically  project 2D image features to explicit BEV feature maps in the 3D space and make predictions based on BEV features \cite{huang2021bevdet,huang2022bevdet4d,li2022bevdepth,li2022bevformer,liu2022bevfusion}. As representative models, BEVDet~\cite{huang2021bevdet}, BEVDepth~\cite{li2022bevdepth} and BEVFusion~\cite{liu2022bevfusion} distribute the 2D features into 3D space according to the estimated depth map, while BEVFormer~\cite{li2022bevformer} adopts cross attention to query BEV features from 2D images. With expressive spatial semantics of BEV, these models achieve the state-of-the-art results on popular benchmarks (\eg, nuScenes~\cite{caesar2020nuscenes}).

\begin{figure*}[t]
     \centering
     \begin{subfigure}[t]{0.25\textwidth}
         \centering
         \includegraphics[width=\textwidth]{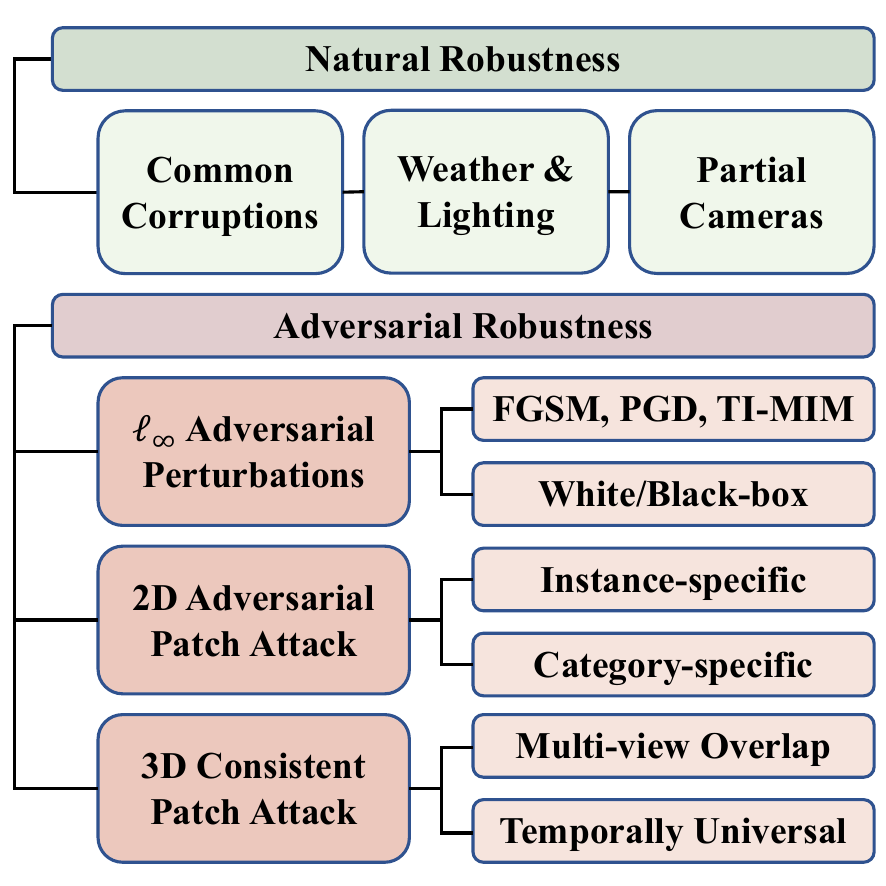}
         \caption{Settings for Robustness Evaluation}
         \label{fig:overview}
     \end{subfigure}
     \hfill
     \begin{subfigure}[t]{0.68\textwidth}
         \centering
         \includegraphics[width=\textwidth]{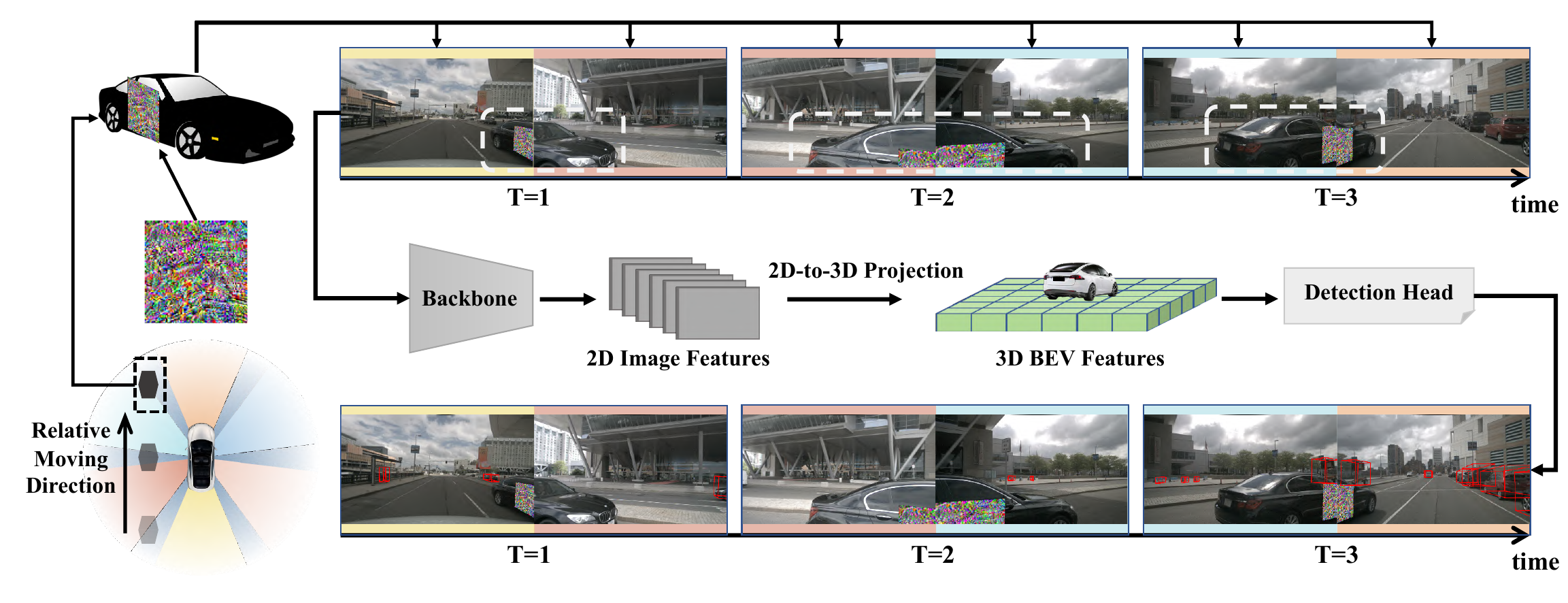}
         \caption{Pipeline of 3D Consistent Patch Attack}
         \label{fig:pipeline}
     \end{subfigure}
\caption{\textbf{Overview. }(a) We measure the natural and adversarial robustness of vision-dependent BEV models in 3D object detection under various settings to thoroughly understand the influence of explicit BEV representations on robustness. (b) By pasting an adversarial patch on a car in the 3D space and projecting it to the 2D images, the generated patch is aligned spatially (across adjacent cameras) and temporally (across continuous frames). The patch cloaks the car in all 3 frames from BEVDepth~\cite{li2022bevdepth}, bringing high safety risks to autonomous driving.}
\vspace{-1ex}
\label{fig:overview_and_pipeline}
\end{figure*}

Despite the excellent performance, these models are still far from practical deployment due to the robustness issues. Previous works have shown that deep learning models are vulnerable to adversarial examples~\cite{szegedy2013intriguing,goodfellow2014explaining}, common corruptions~\cite{hendrycks2018benchmarking}, natural transformations~\cite{engstrom2019exploring,dong2022viewfool}, \etc. The robustness issues rooted in the data-driven deep learning based 3D object detectors can raise severe concerns about the safety and reliability of autonomous driving, making it imperative to evaluate and understand model robustness before being deployed.
As vision-dependent BEV models achieve superior performance and become increasingly prevalent in the field, it is of particular importance to compare their robustness to other models that do not rely on BEV representations, given the inherent trade-off between accuracy and robustness~\cite{su2018robustness,zhang2019theoretically}.

In this paper, we take the first step to systematically analyze and understand the robustness of representative vision-dependent BEV models, by performing thorough experimental evaluations ranging from natural robustness to adversarial robustness as illustrated in~\cref{fig:overview}. We draw several important findings as below:
\begin{itemize}
    \item We first evaluate the natural robustness under common corruptions, various weather and lighting conditions, and partially missing cameras. We find that camera-based BEV models are generally more robust to natural corruptions of images as a result of the rich spatial information carried by BEV representations.
    \item We then evaluate the adversarial robustness under the global $\ell_p$ adversarial perturbations, instance-level and category-level adversarial patches. We observe that BEV models are more vulnerable to adversarial noises, owing to the redundant spatial features represented by BEV based on an in-depth analysis.
    \item Based on the results, we find that camera-LiDAR fusion models have superior performance under all settings due to the aid of multi-modal inputs. Besides, BEVFusion~\cite{liu2022bevfusion} is less robust when both point cloud and image perturbations are imposed.
\end{itemize}

In addition to digital adversarial patches, we propose a novel attack method called \textbf{3D consistent patch attack}. As shown in~\cref{fig:pipeline}, adversarial patches are attached to objects for sptiotemporal consistency in the 3D space. We provide two case studies of 3D consistent patch attack. First, we paste patches on objects falling into the overlap regions of multiple cameras, which are observed in different shapes from different viewpoints. Second, we generate temporally universal patches for objects across a continuous sequence of frames in a certain scene, which is a step further from case one. Both spatial alignment and temporal consistency are considered, which distinguishes 3D object detection for autonomous driving cars from the traditional 2D object detection task.
The conclusions are consistent with those of adversarial robustness above and can inspire more works to guarantee safe autonomous driving.

\begin{table*}[]
\centering
\begin{tabular}{l|c|c|c|c|c}
\hline
            & Modality     & With Transformer       & BEV Repr. & \#Params & mAP / NDS   \\\hline
FCOS3D~\cite{wang2021fcos3d}      & camera-only  & No         & No        & 55M      & 29.8 / 37.7 \\
BEVDet~\cite{huang2021bevdet}      & camera-only  & No         & Yes       & 48M      & 29.2 / 37.2 \\
BEVDepth~\cite{li2022bevdepth}    & camera-only  & No         & Yes       & 53M      & 33.2 / 40.4 \\
DETR3D~\cite{wang2022detr3d}      & camera-only  & Yes & No        & 54M      & 34.7 / 42.2 \\
BEVFormer~\cite{li2022bevformer}   & camera-only  & Yes & Yes       & 60M      & 37.0 / 47.9 \\\hline
TransFusion~\cite{bai2022transfusion} & camera-LiDAR & Yes & No        & 37M      & 67.2 / 70.9 \\
BEVFusion~\cite{liu2022bevfusion}   & camera-LiDAR & Yes & Yes       & 41M      & 68.5 /	71.4 \\\hline
\end{tabular}
\vspace{-1ex}
\caption{\textbf{3D object detection models.} Basic information of different models evaluated in this paper, including input modality, Transformer block, BEV representation, number of parameters and clean detection performance on nuScenes validation set.}
\vspace{-1ex}
\label{tab:model_info}
\end{table*}

\section{Related Work}

\subsection{Vision-Dependent 3D Object Detection}

3D object detection~\cite{zhou2018voxelnet,lang2019pointpillars,philion2020lift,wang2021fcos3d,wang2022detr3d} is important for autonomous driving cars to understand the surrounding objects and therefore safely navigate through the scenes. Camera-based methods~\cite{wang2021fcos3d,wang2022detr3d,liu2020smoke,rukhovich2022imvoxelnet} have been widely studied recently because of its low cost for deployment. Camera inputs are also fused with point clouds in camera-LiDAR fusion models~\cite{bai2022transfusion,chen2022futr3d,yin2021multimodal,vora2020pointpainting}, to capture both rich semantic information in images and spatial information in point clouds. 
Learning from map restoration~\cite{philion2020lift,ng2020bev} that BEV can serve as an effective representation of the environments for 3D perception tasks, methods with BEV representations are proposed for vision-dependent detectors~\cite{huang2021bevdet,huang2022bevdet4d,li2022bevdepth,li2022bevformer,liu2022bevfusion}. They project 2D image features into explicit BEV feature maps in 3D space in order to perform more accurate 3D object detection with expressive spatial information. Specifically, the transformation from 2D to 3D varies across different methods, including projecting 2D features based on depth estimation~\cite{huang2021bevdet,li2022bevdepth,liu2022bevfusion} and querying 2D features from 3D space with cross attention~\cite{li2022bevformer}. Since the perception in auto-driving is performed on a sequence of temporally continuous frames, some methods~\cite{li2022bevformer} take the history BEV features of previous frames into account when detecting in the current frame. The holistic spatial representation by BEV boosts the performance of vision-dependent methods to lead the popular benchmarks~\cite{caesar2020nuscenes,geiger2013vision,sun2020scalability}.

\subsection{Robustness Evaluation of Object Detection}

Robustness has been a crucial issue for deep neural networks (DNN)~\cite{szegedy2013intriguing,goodfellow2014explaining,hendrycks2018benchmarking, chen2021unrestricted}, especially concerning their applications like auto-driving. It has been shown that DNN-based object detectors can be vulnerable to multiple threats including adversarial examples~\cite{xie2017adversarial,wu2020dpattack,zhang2021adversarial,abdelfattah2021towards,chen2021camdar}, common corruptions~\cite{michaelis2019benchmarking,mirza2021robustness,sun2022benchmarking}, \etc, in either 2D or 3D domains. Specifically for 3D object detection, there has been some works evaluating the robustness of LiDAR-based or fusion models~\cite{yu2022benchmarking,albreiki2022robustness,sun2022benchmarking,park2021sensor,mirza2021robustness}. However, the robustness of camera-based 3D object detectors~\cite{zhang2021evaluating}, especially those with BEV representations, has not been fully exploited. This is where our work stands.

 Efforts have been devoted to designing effective adversarial patches~\cite{liu2018dpatch,thys2019fooling,wang2021towards,zhao2020object,huang2021rpattack} for attacking object detectors. Many works take use of Expectation of Transformation (EOT)~\cite{athalye2018synthesizing,lee2019physical} and physically-printable constraints~\cite{thys2019fooling,wang2021towards} to extend 2D-applied patches to successful physical attack, which is a popular topic for the adversarial patch. These methods are further adopted in attacking 3D object detectors~\cite{zhang2021evaluating,park2021sensor}, but is not sufficient for scenarios like auto-driving, where the detection is performed along the time and multi-view cameras are used, which contains plentiful spatial and temporal information. Based on this, we propose 3D consistent patch attack in this paper to apply adversarial patches that are aligned in 3D space.

\section{Preliminary}
\label{sec:setup}

Before evaluating the robustness of vision-dependent BEV detectors and non-BEV detectors in the following sections, we first introduce some notations and experimental setups for better readability. 

Formally, for 3D object detection, we consider the dataset $\mathcal{D}=\cup_{i=1}^NS_i$ containing $N$ scenes, where each scene $S_i=\{(x_t^{(i)},\hat{y}_t^{(i)})|t\in\mathcal{T}\}$ is a sequence of observations $x_t^{(i)}$ and ground-truth labels $\hat{y}_t^{(i)}$ at $T$ continuous timestamps $\mathcal{T}=\{0,\cdots,{T-1}\}$. $x\in[0,255]^{N_c\times 3\times H\times W}$ represents $N_c$ views of RGB images for camera-based models and $x$ can also include point clouds data for fusion models. For a 3D object detector, we denote it as $y=f_\theta(x)$ in general, which is trained by an objective function $\mathcal{L}$.
In the following context, we will use these notations by default unless otherwise stated.

For experimental evaluation, we use the validation set of nuScenes~\cite{caesar2020nuscenes} to study 7 modern 3D object detectors as shown in~\cref{tab:model_info}. There are 4 BEV models and 3 non-BEV counterparts. 
Among them, FCOS3D~\cite{wang2021fcos3d}, BEVDet~\cite{huang2021bevdet} and BEVDepth~\cite{li2022bevdepth} are CNN-based, while DETR3D~\cite{wang2022detr3d}, BEVFormer~\cite{li2022bevformer}, TransFusion~\cite{bai2022transfusion} and BEVFusion~\cite{liu2022bevfusion} have Transformer blocks in their architectures. The last two are fusion models and have better performance compared to the aforementioned camera-only models. Models for comparison are chosen to have similar numbers of parameters, since larger models tend to have better robustness~\cite{bhojanapalli2021understanding}. We further rule out additional impacts from training techniques like data augmentation. The detailed strategies for the chosen models are listed in~\cref{sec:appendix_training} and there is an insignificant difference within the three comparing sub-groups. Though BEVDet and BEVDetpth take in images with lower resolution, but our experiments in~\cref{sec:appendix_resolution} show that the conclusions are consistent and the influence is nonessential.  For evaluation metrics, we adopt mean Average Precision (mAP) and nuScenes Detection Score (NDS) which is a composite indicator of a set of True Positive metrics along with mAP.

\begin{table*}[h]
\centering
  \resizebox{\linewidth}{!}{
\begin{tabular}{l|ccc|cccc|cccccc}
\hline
            & \multicolumn{3}{c|}{Noise}    & \multicolumn{4}{c|}{Blur}              & \multicolumn{5}{c}{Digital}                         \\\cline{2-4}\cline{5-8}\cline{9-13}
            & \small{Gaussian} & \small{Shot}    & \small{Impulse} & \small{Defocus} & \small{Glass}   & \small{Motion}  & \small{Zoom}    & \small{Brightness} & \small{Contrast} & \small{Elastic} & \small{Pixel}   & \small{JPEG}    \\\hline
FCOS3D      &  2.7/11.5         &  3.0/11.9        &  3.0/11.9        &  7.0/18.9        &  7.0/19.7        &  7.4/18.3        &  0.3/1.8        &  15.1/26.4          &  11.4/24.8        &   18.9/30.3      &  11.3/24.0       &   11.2/24.4      \\
BEVDet      &  3.3/10.6         &  4.1/12.4        &  2.6/8.5         &  12.8/24.2       &  18.0/28.1       &  14.5/25.7       &   1.2/4.6        &  21.1/29.3          &  10.6/22.1        &   28.9/37.1      &  28.2/36.3       &   17.3/27.5      \\
BEVDepth    &  4.7/11.4         &  6.1/15.4        &  4.6/11.2        &  \textbf{20.1}/30.8       &  \textbf{22.8}/32.2       &  19.8/30.7       &  1.8/7.9        &  23.6/32.8          &  15.8/25.1        &   32.6/40.1      &  31.6/39.3       &   23.8/33.6      \\
DETR3D      &  15.7/27.8        &  17.0/28.5       &  16.2/28.0       &  20.0/30.3       &  16.6/28.2       &  18.8/29.8       &   2.4/11.8       &  33.5/41.6          &  29.1/38.3        &   32.5/40.4      &  28.6/37.7       &   25.8/35.4      \\
BEVFormer   &  \textbf{16.5}/\textbf{32.6}        &  \textbf{18.3}/\textbf{33.9}       &  \textbf{16.8}/\textbf{32.7}       &  19.8/\textbf{35.6}       &  21.0/\textbf{35.7}       &  \textbf{24.2}/\textbf{38.9}       &   \textbf{2.7}/\textbf{14.2}       &  \textbf{35.0}/\textbf{46.7}          &  \textbf{29.2}/\textbf{42.7}        &   \textbf{36.0}/\textbf{47.0}      &  \textbf{33.7}/\textbf{45.9}       &   \textbf{30.2}/\textbf{43.5}      \\\hline
TransFusion &  \textbf{65.9}/\textbf{70.2}        &  \textbf{66.0}/\textbf{70.3}       &  \textbf{65.8}/\textbf{70.2}       &  \textbf{66.5}/\textbf{70.5}       &  66.5/\textbf{70.5}       &  66.4/\textbf{70.5 }      &   \textbf{65.5}/\textbf{70.0}      &  \textbf{67.1}/\textbf{70.8}          &  \textbf{66.3}/\textbf{70.4}        &   67.2/70.9      &  67.1/70.8       &   \textbf{66.9}/\textbf{70.7}      \\
BEVFusion   &  63.4/68.7        &  63.4/68.7       &  63.4/68.7       &  66.1/70.2       &  \textbf{66.6}/70.4       &  \textbf{66.6}/70.4       &   61.9/67.7      &  66.8/70.4          &  66.0/70.0        &   \textbf{68.4}/\textbf{71.4}      &  \textbf{68.3}/\textbf{71.3}       &   66.7/70.6      \\
\hline
\end{tabular}}
\vspace{-2ex}
\caption{\textbf{Common Corruptions. }mAP/NDS of vision-dependent models under 12 common corruptions at level 3 as in ImageNet-C~\cite{hendrycks2018benchmarking}.}
\vspace{-3ex}
\label{tab:common_corruption}
\end{table*}

\begin{table}[]
\centering
\small
\resizebox{\linewidth}{!}{
\begin{tabular}{l|cccc}
\hline
            &  Day         & Night          & Sunny          & Rainy \\\hline
            
FCOS3D      &  30.1/37.9   &   13.9/22.6    &   29.5/37.6    &  30.1/37.6     \\
BEVDet      &  30.2/38.1   &   12.0/21.7    &   29.4/36.5    &  30.1/43.2     \\
BEVDepth    &  33.7/41.0   &   13.2/22.9    &   33.1/39.4    &  33.3/44.7     \\
DETR3D      &  35.0/42.7   &   16.0/23.0    &   34.3/41.0    &  36.1/47.5     \\
BEVFormer   &  \textbf{37.2}/\textbf{48.1}   &   \textbf{20.1}/\textbf{27.3}    &   \textbf{36.6}/\textbf{47.0}    &  \textbf{38.3}/\textbf{50.8}     \\\hline
TransFusion &  67.3/71.0   &   39.8/44.7    &   67.0/70.5    &  67.5/71.9     \\
BEVFusion   &  \textbf{68.5}/\textbf{71.5}   &   \textbf{43.9}/\textbf{46.7}    &   \textbf{68.4}/\textbf{71.3}    &  \textbf{69.5}/\textbf{72.2}    \\\hline
\end{tabular}
}
\vspace{-2ex}
\caption{\textbf{Weather \& Lighting.} mAP/NDS of vision-dependent models under different weather and lighting conditions. 
}
\vspace{-3ex}
\label{weather}
\end{table}

\section{Natural Robustness}
\label{sec:natural}

In this section, we first evaluate the natural robustness of vision-dependent BEV models under common corruptions, different weather and lighting conditions, and partial cameras. The natural robustness of a model is a critical issue in real world applications~\cite{hendrycks2018benchmarking,hendrycks2021natural}, which indicates its reliability, stability and consistency.

\subsection{Common Visual Corruptions}

\textbf{Motivation and Setting.} Various forms of vision corruptions like noise, blur and digital distortions could happen in the scenario of autonomous driving due to high speed or malfunctioning sensors~\cite{hendrycks2018benchmarking}, presenting potential safety risks. Four main categories of common corruptions are proposed in~\cite{hendrycks2018benchmarking} to evaluate the robustness of image classifiers. Similarly, we analyze the performance of vision-dependent BEV detectors under 12 different corruptions, including noise, blur and digital distortions at level 3.

\textbf{Results. }The model performance in terms of mAP and NDS  under various corruptions is shown in~\cref{tab:common_corruption}. We make the observations below. 
First, the corruptions imposed on the images barely influence the detection of TransFusion and BEVFusion. The complementary information contained in point clouds makes the fusion models robust under image corruptions. 
Second, methods with Transformer blocks are more robust to natural corruptions compared to FCOS3D, BEVDet and BEVDetpth, which are CNN-based. This finding is consistent with the conclusions in previous studies that claim the better robustness of Transformers in image classification~\cite{bhojanapalli2021understanding,naseer2021intriguing}. 
Third, besides the negligible difference between fusion models, there are significant advantages of camera-only BEV models compared to non-BEV methods, especially for corruptions of blur and digital distortions. This indicates that BEV models have better natural robustness to image corruptions in general. The holistic modeling of spatial features by BEV representations fusing the multi-view information contributes to this superiority.

\begin{figure}[t]
     \centering
     \begin{subfigure}[t]{0.32\linewidth}
         \centering
         \includegraphics[width=\linewidth]{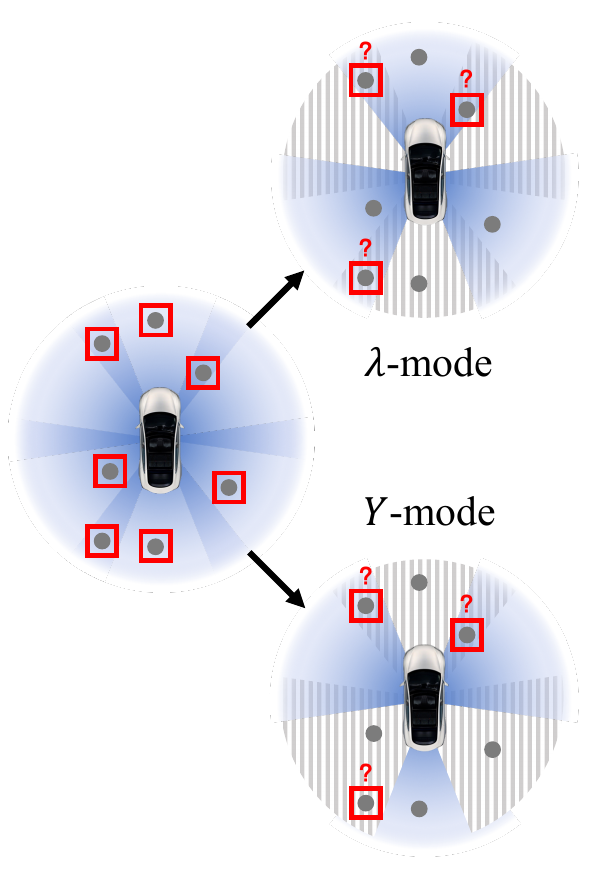}
         \caption{Two modes}
         \label{fig:partial_note}
     \end{subfigure}
     \begin{subfigure}[t]{0.64\linewidth}
         \centering
         \includegraphics[width=\linewidth]{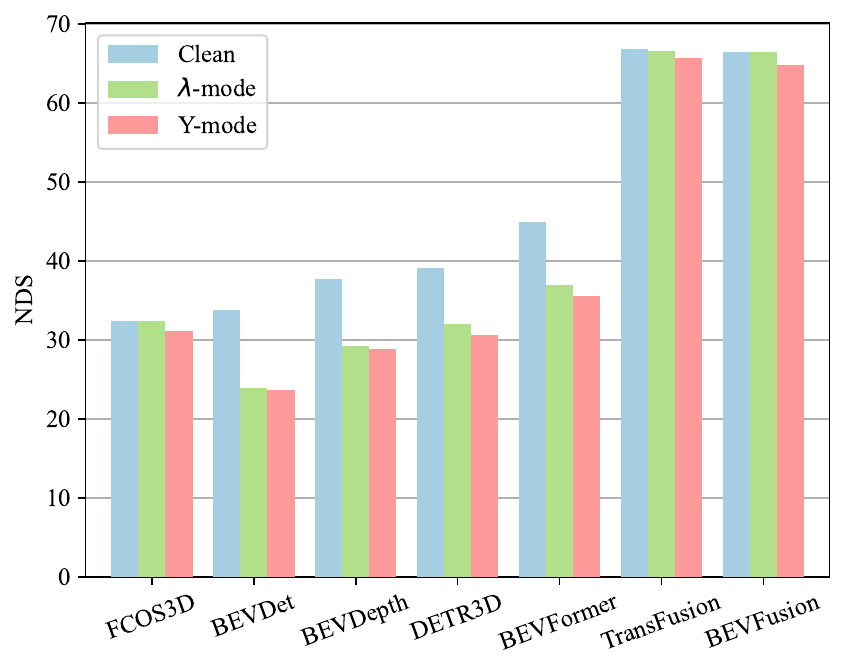}
         \caption{Results of NDS}
         \label{fig:partial_nds}
     \end{subfigure}
        \vspace{-1ex}
        \caption{\textbf{Partial Cameras. }We consider two modes of partially missing cameras, $\lambda$-mode and $Y$-mode, where 3 nonadjacent cameras are masked out and only objects falling into the multi-view overlap regions are focused, as depicted in (a). The performance with partial cameras in terms of NDS respectively across different models is shown in (b).}
        \vspace{-4ex}
        \label{fig:partial_camera}
\end{figure}

\begin{figure*}[t]
     \centering
     \includegraphics[width=0.99\linewidth]{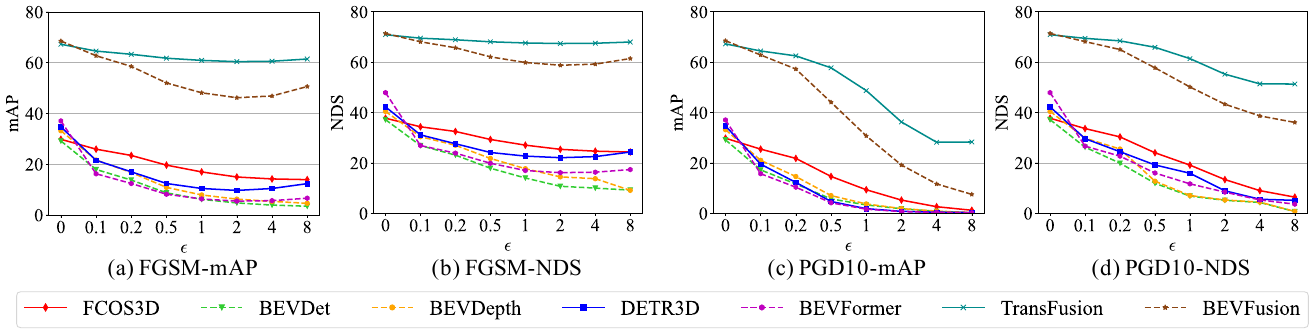}
     \vspace{-2ex}        
     \caption{ \textbf{$\ell_\infty$ Image Perturbations.}
         Results of $\ell_\infty$ adversarial perturbations generated by FGSM~\cite{goodfellow2014explaining} and PGD10~\cite{madry2018towards} with different $\epsilon$ ranging from $0$ to $8$.}
    \vspace{-4ex}
     \label{fig:lp-attack}
\end{figure*}

\subsection{Weather and Lighting Conditions}

\textbf{Motivation and Setting. }It is common to encounter different weather and lighting conditions for autonomous driving, which can cause the loss of image quality as well as a shift of data distribution~\cite{subbaswamy2021evaluating}, and further lead to worse performance. Following the setting in~\cite{liu2022bevfusion}, we analyze the performance of models under different weather and lighting conditions by dividing the dataset into 4 subsets, including Day, Night, Sunny and Rainy, according to the annotations in nuScenes~\cite{caesar2020nuscenes}.

\textbf{Results. }The results of different models under various weather and lighting conditions are shown in~\cref{weather}. Generally, though poor lighting condition (Night) brings great difficulties to the detection, BEV models have better performance on 4 subsets, which is consistent to clean performance on the complete dataset.

\subsection{Partial Cameras}
\label{sec:partial_cam}


\textbf{Motivation and Setting. }The multi-view detection with BEV features enables the global perception of the surrounding environment, but may suffer from incomplete data when some cameras break down. To study model robustness in this case, we consider objects falling into the overlap regions of adjacent cameras~\cite{wang2022detr3d}, \ie, they are captured by multiple cameras, thus the detection of objects is feasible in the absence of one view due to the global representation of BEV. As shown in~\cref{fig:partial_note}, we mask out 3 non-adjacent cameras out of the 6 multi-view cameras in nuScenes, resulting in 2 settings---$\lambda$-mode and $Y$-mode. 

\textbf{Results. }The corresponding NDS is plotted in~\cref{fig:partial_nds}. With partial cameras, the detection performance decreases to different extents for all models. Fusion models with partial camera inputs and complete LiDAR inputs are slightly affected. FCOS3D, a monocular detector, exhibits a negligible performance drop since it conducts detection on individual images. BEVFormer outperforms DETR3D in terms of two metrics, which indicates that holistic BEV representations from vision input contribute to the spatial modeling of multi-view overlap regions. Meanwhile, the metrics in $\lambda$-mode are universally higher than that in $Y$-mode. This suggests that there is a bias in multi-view detection among these cameras and the detection in the front is better. 

\section{Adversarial Robustness}
\label{sec:adversarial_robustness}

Adversarial examples~\cite{szegedy2013intriguing,goodfellow2014explaining} are inputs with carefully crafted noises by an adversary to mislead the models. They help to measure the worst-case performance especially under malicious attacks.  In this section, we conduct a series of classic untargeted adversarial attacks to assess the adversarial robustness of BEV methods, by maximizing the training objectives of the detectors with adversarial noises. The formulations of adversarial attacks in this section are introduced in~\cref{sec:appendix-formulations} as a background guide.

\subsection{$\ell_\infty$ Adversarial Perturbations}
\label{sec:lp-attack}
\textbf{Motivation. }The pioneering methods~\cite{szegedy2013intriguing,goodfellow2014explaining,dong2018boosting} first propose to add small, human-imperceptible noises to
legitimate examples to induce wrong model outputs. This attack scheme allows us to primarily understand the behaviors of models when encountering small malicious perturbations in the digital space. The scale of the perturbations can be measured by the $\ell_p$ distance between the adversarial example and the clean input and the $\ell_\infty$ norm is used here.

\textbf{Settings. }We adopt two typical gradient-based attack methods---Fast Gradient Sign Method (FGSM)~\cite{goodfellow2014explaining} and Projected Gradient Descent (PGD)~\cite{madry2018towards}, to generate $\ell_\infty$ adversarial perturbations for each model. The budget of perturbation $\epsilon$ varies from $0$ (clean) to $8$ under the $\ell_\infty$ norm. The number of iterations for PGD is set to $10$ as PGD10.

\begin{figure*}[t]
     \centering
     \begin{subfigure}[t]{0.45\linewidth}
         \centering
         \includegraphics[width=\linewidth]{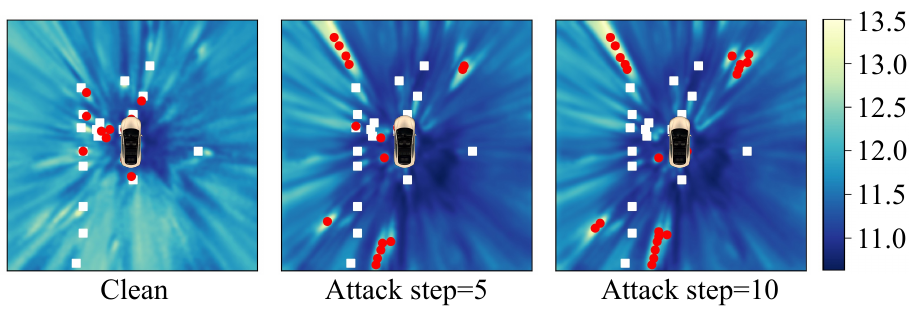}
         \caption{Visualized BEV feature map with predictions~(red dots) and ground truth~(white squares) of BEVFormer at different attack steps.}
         \label{fig:bev-map}
     \end{subfigure}
     \hspace{1ex}
     \begin{subfigure}[t]{0.45\linewidth}
         \centering
         \includegraphics[width=0.84\linewidth]{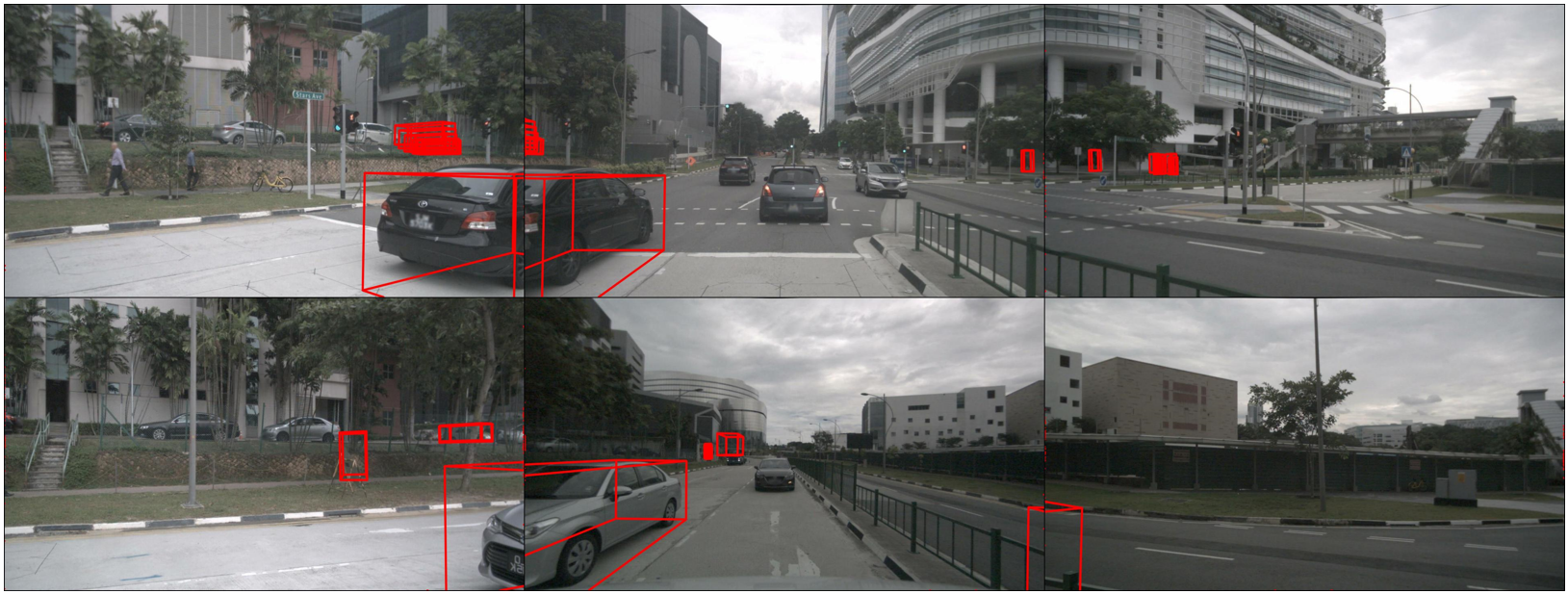}
         \caption{Visualized predictions~(red bounding boxes) of BEVFormer under PGD10 attack.}
         \label{fig:bev-map-prediction}
     \end{subfigure}
     \vspace{-2ex}
     \caption{\textbf{Qualitative Analysis of BEV model. }Visualization of BEV feature map and predictions under PGD10 attack.}
     \vspace{-3ex}
     \label{fig:visual_bev}
\end{figure*}

\textbf{Results. }We show the curves of mAP and NDS of different models under attacks along with $\epsilon$ in~\cref{fig:lp-attack}. Overall, all models are affected by the adversarial noises to varying degrees and the impact grows with more iterations and larger $\epsilon$. 
First, the changes in the performance of fusion models are less than camera-only models. 
For instance, the NDS of BEVFusion at $\epsilon=8$ under FGSM drops 10.0 compared to the clean performance, while for BEVFormer, it drops 30.6. 
This implies that fusion models are more robust to adversarial perturbations imposed on images due to the multi-modality inputs including point clouds, which is similar to the conclusion in~\cref{sec:natural}. 
Second, though BEV models have better performance on clean images, there is a consistent phenomenon that the impacts on the performance of BEV models are more severe and their resulted mAP and NDS are lower than their non-BEV counterparts. A significant result is that BEVFormer achieves the highest mAP of 37.0 on clean images and declines to one of the lowest mAP of 6.3 with PGD at $\epsilon=1$ among camera-only methods. This leads to the conclusion that although BEV models have better performance in 3D object detection, they have inferior adversarial robustness. 
Besides, we notice that the performance of some models rises a little after the $\epsilon$ comes 4 and 8 in FGSM attack.
We attribute the unexpected rise to the highly-nonlinear loss landscape of 3D detectors.
With a large step size, generated perturbations could fail to cross the decision boundary and therefore result in inferior performance.

\textbf{Analysis. }To further pinpoint the underlying reason why BEV models are vulnerable to $\ell_\infty$ perturbations, we conduct some qualitative analysis by comparing the behaviors between DETR3D~\cite{wang2022detr3d} and BEVFormer~\cite{li2022bevformer}, which share the same backbone of ResNet101 and use Transformer blocks to perform detection. We statistically compute the changes on image features extracted by ResNet101 in terms of normalized mean square error (NMSE), when the models are under attack by PGD10 with $\epsilon=1$. The NMSE of DETR3D on 200 samples is 4.7488~($\sigma$=0.2863) while that of BEVFormer is 4.8042~($\sigma$=0.2675), which indicates that for two different models, the influence on feature extraction from PGD10 is similar. Thus, the difference in adversarial robustness should come from stages after feature extraction, \ie, BEV modeling for BEVFormer. We analyze the BEV feature maps under attack qualitatively to study why BEV models are vulnerable. One example of visualization is shown in~\cref{fig:visual_bev}. We find that adversarial perturbations lead to greater activations in areas without objects and further generate numerous false positives. Since BEV representations model the whole 3D space, there are redundant features for the adversarial attack to corrupt, which leads to worse robustness of BEV models.

\begin{figure}
     \centering
     \includegraphics[width=0.99\linewidth]{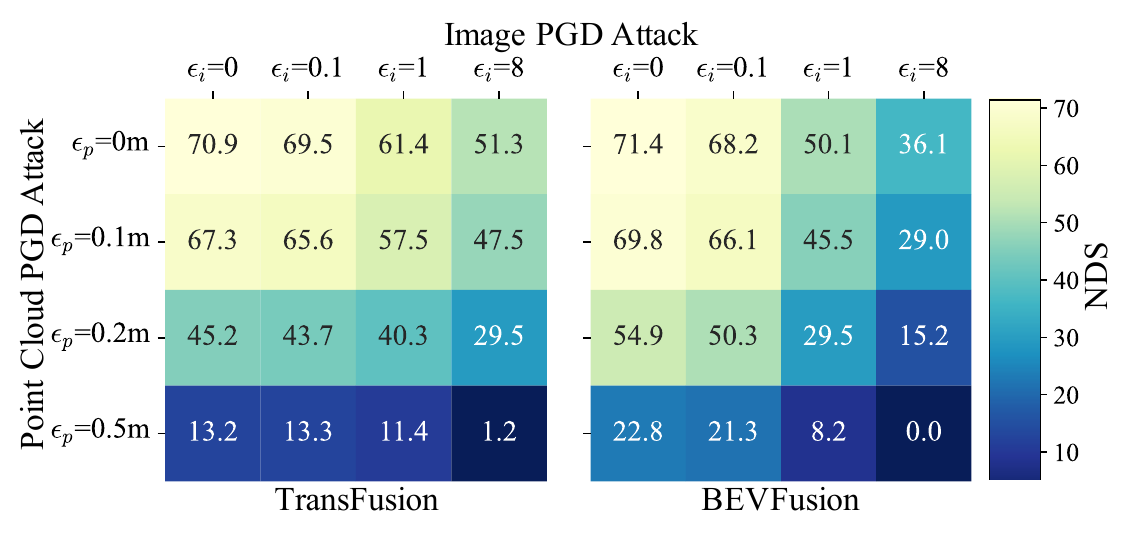}
     \vspace{-2ex}
     \caption{\textbf{$\ell_\infty$ Perturbations for Fusion Models. }NDS of Camera-LiDAR fusion models under different adversarial perturbations added to images and point clouds.}
     \vspace{-3ex}
     \label{fig:pointimg_NDS}
\end{figure}

\textbf{Point Cloud Perturbations. }Although being less affected by image perturbations, fusion models are also exposed to adversarial noises to both images and point clouds. We still adopt PGD10 with a $4\times 4$ design of $\ell_\infty$ budgets for image and point cloud perturbations, as shown in~\cref{fig:pointimg_NDS}. We find that there is a trade-off between robustness to image perturbations and point cloud perturbations. BEVFusion is less robust than TransFusion when only image perturbations are imposed, but more robust with only point cloud perturbations. Also, when two kinds of perturbations are both applied, BEVFusion tends to be more vulnerable. The reason is that BEVFusion relies more on image inputs in its pipeline, \ie, it aggregates image and point cloud features to holistic BEV representations to generate object proposals, while TransFusion only uses point cloud features in its early proposal stage. Therefore, larger image perturbations will further decrease the adversarial robustness to point cloud perturbations for BEVFusion. This can also account for the slightly inferior performance of BEVFusion compared to TransFusion under previous image corruptions.

\textbf{Transfer Attack. }Adversarial examples are shown to have cross-model transferability~\cite{dong2018boosting,liu2016delving,papernot2016transferability}, which enables practical black-box attacks. We further study the transferability of adversarial examples between different 3D detection models. We find that the transferability is better across models with similar architectures, while the effect of BEV features is insignificant. Details are listed in~\cref{sec:appendix-linf}.

\subsection{Instance-specific Adversarial Patches}
\label{sec:instance-patch}

\textbf{Motivation. }Adversarial examples with $\ell_p$ perturbations are infeasible for physical attack due to pixel-wise modifications. Patch attack with localized and visible perturbations is therefore studied~\cite{brown2017adversarial,liu2018dpatch,wu2020dpattack,thys2019fooling,zhang2021adversarial,song2018physical}. The adversarial patch is often a square pattern masked over the victim object for attack in object detection. We first consider the instance-specific adversarial patches, \ie, for each frame, we generate adversarial patches for every single object under each view respectively.

\textbf{Settings. }We set the locations of 2D patches at the 2D positions projected from the centers of 3D bounding boxes, and their sizes are proportional to the sizes of the projected 2D bounding boxes at different ratios, ranging from 1\% to 10\%. Patch colors for each frame are optimized in 20 steps by Adam~\cite{kingma2014adam} optimizer with learning rate of $0.1$.

\begin{table}[t]
\centering
\small
\resizebox{\linewidth}{!}{
\begin{tabular}{l|cccc}
\hline
    Patch Size        & 1\%                          & 2\%                           & 5\%                           & 10\%                          \\\hline
FCOS3D      & \textbf{18.3}/\textbf{27.2}                   & \textbf{14.6}/23.2              & \textbf{8.4}/16.2                  & \textbf{3.9}/11.8                  \\
BEVDet      & 10.1/18.8                   & 6.6/13.9                  & 3.0/7.9                  & 1.1/2.9                  \\
BEVDepth    & 13.3/21.7                  & 9.2/18.6                  & 4.3/9.1                  & 1.7/7.3                  \\
DETR3D      & 16.8/ 26.9                  & 12.1/\textbf{23.3}                  & 6.0/\textbf{16.8}                 & 2.4/\textbf{12.5 }                 \\
BEVFormer   & 12.7/24.5                  & 9.4/20.2                  & 4.9/15.9                   & 2.1/11.0                  \\\hline
TransFusion & \textbf{62.5}/\textbf{68.3}                  & \textbf{60.5}/\textbf{67.2}                  & \textbf{55.6}/\textbf{64.7}                  & \textbf{46.5}/\textbf{60.1}                  \\
BEVFusion   & 57.0/64.9                  & 50.9/61.5                  & 39.1/55.0                 & 29.1/49.5                 \\\hline
\end{tabular}
}
\vspace{-2ex}
\caption{
\textbf{Instance-Specific Adversarial Patches. }mAP/NDS of vision-dependent models with instance-specific adversarial patches of different size ratios.}
\vspace{-3ex}
\label{tab:instance-patch}
\end{table}

\textbf{Results. }The results of instance-specific patch attack are shown in~\cref{tab:instance-patch}. With the size ratio increasing, the performance of all models keeps declining. In general, the results show the similar trends with $\ell_\infty$ adversarial perturbations. 
Fusion models still demonstrate better robustness to the local adversarial noises. 
Also, the performance of all BEV methods is worse than non-BEV methods and the gaps are significant especially for camera-only models without Transformer block (BEVDet, BEVDepth \vs FCOS3D) and fusion models (BEVFusion \vs TransFusion). Since the frame-by-frame attack setting is only distinguished from $\ell_\infty$ perturbations with localized noises, it is reasonable that BEV models have the similar conclusion of worse adversarial robustness and the underlying reasons are similar.

\subsection{Category-specific Adversarial Patches}
\label{sec:class-patch}
\textbf{Motivation. }Furthermore, universal attack is often considered in the context of adversarial patch~\cite{brown2017adversarial,chou2020sentinet,liu2018dpatch,zolfi2021translucent}, because it's more feasible to perform physical attack when the objects and models are not certain. We consider category-specific adversarial patches, \ie, for each category of objects in the dataset, we generate a adversarial patch for it.

\textbf{Settings. }The patch location is the same as it in \cref{sec:instance-patch}. We pre-define a $100\times 100$ patch for each category and resize them to the proper size when applying them to the object. The patch colors are optimized across the dataset for 3 epochs by the Adam optimizer with learning rate of $0.01$.

\textbf{Results. }With similar settings but universal attack, we draw a similar finding that fusion models have better robustness and BEV models have worse robustness from the results in~\cref{tab:category-patch}. However, an interesting phenomenon occurs that BEVFormer has better performance than DETR3D when applied with category-level adversarial patches. A possible explanation is that BEVFormer predicts based on history BEV features, which neutralize the computation of the expectation of gradients over the dataset and make it harder to conduct the universal attack.

\begin{table}[]
\centering
\small
\resizebox{\linewidth}{!}{
\begin{tabular}{l|cccc}
\hline
Patch Size  & 1\%       & 2\%       & 5\%         & 10\% \\\hline
FCOS3D      & 24.9/28.8 & 22.0/26.7 & 14.5/22.2   & \textbf{14.5}/22.2      \\
BEVDet      & 16.1/27.8 & 9.3/21.9  & 3.6/15.4    & 2.0/8.2     \\
BEVDepth    & 21.9/31.9 & 16.3/26.4 & 8.3/20.4    & 2.4/8.5     \\
DETR3D      & 31.3/39.4 & 26.1/36.8 & 15.5/29.5   & 7.3/22.8     \\
BEVFormer   & \textbf{33.2}/\textbf{45.0} & \textbf{30.9}/\textbf{43.4} & \textbf{19.8}/\textbf{35.9}   & 8.1/\textbf{24.0}     \\\hline
TransFusion & 66.6/\textbf{70.6} & 66.3/\textbf{70.4} & \textbf{65.5}/\textbf{70.0}   & \textbf{65.0}/\textbf{69.7}     \\
BEVFusion   & \textbf{67.2}/\textbf{70.6} & \textbf{66.4}/70.2 & 64.3/69.2   & 61.4/67.7     \\\hline    
\end{tabular}
}
\vspace{-2ex}
\caption{\textbf{Category-Specific Adversarial Patches. }mAP/NDS of vision-dependent models with category-specific adversarial patches of different size ratios.}
\vspace{-3ex}
\label{tab:category-patch}
\end{table}

\section{3D Consistent Patch Attack}

Adversarial attacks in~\cref{sec:adversarial_robustness} are mostly proposed in 2D vision tasks of image classification and object detection. Considering that 3D object detection in autonomous driving involves multi-view cameras and continuous frames, we propose a \textbf{3D consistent patch attack}, which applies adversarial patches on objects in the real 3D space. This attack considers the 3D consistency and alignment, which is more appropriate for autonomous driving. 

\subsection{Method}

Rather than patch attack in digital space as in~\cref{sec:adversarial_robustness}, where the process of applying a patch is a pixel-wise multiplication between masking matrix and 2D images, 3D consistent patch attack requires an applying function $\mathcal{A}$ to project the adversarial patches in the 3D space to 2D perspective views of cameras.

We then introduce the process of $\mathcal{A}$ formally. For an object in a single frame, we paste the corresponding adversarial patch, whose size is $H_p\times W_p$ in the real world, to an appointed 3D position of the object. Given the information of ground-truth 3D bounding box, the 3D positions of the corners of the pasted patch can be computed, which are located in the LiDAR coordinate system and represented by $(p_1,p_2,p_3,p_4)$, where $p_i\in\mathbb{R}^3$. For each camera view, there is a projection matrix $\mathcal{M}_{3d-2d}\in\mathbb{R}^{4\times 4}$ according to the camera intrinsic parameters and a 3D point $p=(x_p,y_p,z_p)^\top$ in LiDAR coordinates can be projected to point $p'\in\mathbb{R}^2$ on 2D image plane following
\begin{equation}
    \begin{bmatrix}
        x_c\\y_c\\z_c\\1
    \end{bmatrix}
    =\mathcal{M}_{3d-2d}\begin{bmatrix}
        x_p\\y_p\\z_p\\1
    \end{bmatrix},\; p'=(x_c/z_c, y_c/z_c)^\top.
\end{equation}
Then, we digitally apply the patch to the quadrangle defined by the projected corners $(p'_1,p'_2,p'_3,p'_4)$ with perspective transformation. A vector of projection coefficients $[a,b,c,d,e,f,g,h]^\top\in\mathbb{R}^8$ is used and the pixel $(h_t, w_t
)$ inside the target quadrangle are projected back to the location $(h_s,w_s)$ on the source $H_p\times W_p$ patch following
\begin{equation}
\begin{bmatrix}
    h_s\\w_s
\end{bmatrix}
=\begin{bmatrix}
    a\cdot h_t+b\cdot w_t+c/g\cdot h_t+h\cdot w_t+1\\d\cdot h_t+e\cdot w_t+f/g\cdot h_t+h\cdot w_t+1
\end{bmatrix},
\end{equation}
where the coefficients can be solved with the corresponding corner coordinates. Then, the pixel color at $(h_t, w_t)$ on the 2D image can be interpolated across the neighbor pixels around $(h_s,w_s)$ on the patch. The process is differentiable and the original adversarial patches can be optimized in a similar way as 2D patches in~\cref{sec:instance-patch} and~\cref{sec:class-patch}.

In the following, we consider two cases that can apply the proposed 3D consistent patches and line with the distinguished properties of 3D object detection in auto-driving.

\subsection{Multi-view Patch Attack}

\textbf{Motivation and Formulation. }Multi-view detection is a unique property of autonomous driving. An adversarial patch in 3D space can be captured from different angles and thus deforms to different shapes on 2D images. In this case, 3D patches are attached to the objects in multi-view overlap regions.  The patch has different ratios of size similar to those in~\cref{sec:instance-patch}, and its plane is vertical to the direction pointing the car. The multi-view patch attack is conducted frame by frame and formalized as
\begin{equation}
    \max_{\mathbf{p}=\{p_1,\cdots,p_{N_{ol}}\}} \mathcal{L}(f_\theta(\mathcal{A}(x,\mathbf{p})),\hat{y}),
\end{equation}
where $\mathbf{p}$ is the set of 3D adversarial patches for the $N_{ol}$ objects in the overlap regions in one specific frame, which are applied to the model inputs with $\mathcal{A}$.

\textbf{Results. }The performance is shown in~\cref{tab:overlap}, which is only evaluated on objects in overlap regions. Fusion models are still superior to camera-only methods either in clean detection or with 3D consistent patches. BEV models demonstrate better or closed clean detection of overlap regions, but the performance is worse than non-BEV models under attack. Though the form of adversarial noises has changed to spatially aligned 3D consistent patches, the attack is still white-box and performed frame by frame. Therefore, the conclusions of adversarial robustness for both fusion models and BEV models in~\cref{sec:lp-attack} and~\cref{sec:instance-patch} still hold.

\subsection{Temporally Universal Patch Attack}

\begin{table}[]
\centering
\small
\resizebox{0.9\linewidth}{!}{
\begin{tabular}{l|ccc|ccc}
\hline
Case     & \multicolumn{3}{c|}{Multi-view Patch} &   \multicolumn{3}{c}{Temp. Univ. Patch} \\\hline
Patch Size     & 0\%           & 5\%           & 10\%          & 0\%           & 5\%           & 10\% \\\hline
FCOS3D         &  32.5         & \textbf{22.2} & \textbf{17.5} &  37.7         & 20.6          & 15.3     \\
BEVDet         &  33.8         & 11.1          & 6.2           &  37.2         & 12.7          & 5.9   \\
BEVDepth       &  37.7         & 11.8          & 8.8           &  40.4         & 17.7          & 8.7   \\
DETR3D         &  39.1         & 15.2          & 10.3          &  42.2         & 28.3          & 23.2    \\
BEVFormer      &  45.0         & 14.8          & 9.6           &  47.9         & \textbf{35.0} & \textbf{29.0}    \\\hline
TransFusion    & \textbf{66.8} & \textbf{63.7} & \textbf{62.8} &  70.9         & \textbf{68.0} & \textbf{66.4}     \\
BEVFusion      &  66.4         & 54.2          & 50.8          & \textbf{71.4} & 63.9          & 60.7   \\\hline
\end{tabular}}
\vspace{-2ex}
\caption{\textbf{3D Consistent Patch Attack. }NDS of vision-dependent models with 3D consistent patches in the cases of Multi-view Patch Attack and Temporally Universal Patch Attack respectively. 0\% denotes clean images.}
\vspace{-3ex}
\label{tab:overlap}
\end{table}

\textbf{Motivation and Formulation. }The frame-to-frame temporal continuity is another distinguished property in the auto-driving system. Most attack settings considered in previous sections are for individual frame and the temporal consistency is not taken into account. We propose a setting where adversarial patches are applied to the surrounding surfaces of every object appeared in a sequence of frames and these patches are spatial-temporally consistent. For a sequence of observations in the $i$-th scene $S_i$, the attack is formalized as
\begin{equation}
    \max_{\mathbf{p}^{(i)}=\{p^{(i)}_1,\cdots,p^{(i)}_{N_{i}}\}} \mathbb{E}_{t\sim\mathcal{T}}[\mathcal{L}(f_\theta(\mathcal{A}(x^{(i)}_t,\mathbf{p}^{(i)})),\hat{y}_t^{(i)})],
\end{equation}
where $\mathbf{p}^{(i)}$ is the set of 3D adversarial patches for the $N_{i}$ objects appeared in scene $S_i$ and remains consistent across the timeline to perform the temporally universal attack. 

\textbf{Results. }We show the results in~\cref{tab:overlap}. Despite the impressive performance of fusion models and better adversarial robustness of FCOS3D and TransFusion, we find that BEVFormer outperforms DETR3D by at least 2.2 in mAP and 5.8 in NDS, demonstrating better robustness in this case. This is similar to the category-specific patch attack. Both settings require universal noises across instances in a category or frames in a sequence, and therefore the better robustness of BEVFormer should also come from the temporal dependency of history BEV features.

\section{Discussion and Conclusion}
In this paper, we delve into the robustness of vision-dependent BEV models in 3D object detection~\cite{li2022bevdepth,huang2021bevdet,li2022bevformer,liu2022bevfusion}.
We have studied the natural and adversarial robustness of 7 popular models under different settings, and propose a novel method of 3D consistent patch attack specifically designed for the scenario of autonomous driving. Various phenomena are observed and many conclusions are drawn from them, some of which confirm the previous observations while others are inspiring for the design of BEV models in 3D object detection:
\begin{itemize}
    \item The holistic representations of BEV with rich spatial and semantic features improve the detection accuracy as well as the natural robustness of 3D detectors to common corruptions of images and other forms of disturbance like partially missing cameras.
    \item Under intentional adversarial attacks, BEV models tend to be more vulnerable to the adversarial noises on images (\eg, $\ell_\infty$ perturbations and adversarial patches), which results from the extra activation on the BEV feature maps as analyzed on BEVFormer.
    \item Fusion models generally have better performance and robustness than camera-only methods due to the multi-modality, and BEVFusion is more vulnerable to adversarial attacks when perturbations on both images and point clouds are applied because of its greater dependence on vision inputs.
    \item Consistent with~\cite{bhojanapalli2021understanding,bai2021transformers} in image classification, Transformer improves the performance and natural robustness of 3D detectors, while they are not more adversarially robust than pure CNN models. 
    \item As shown by results of BEVFormer in category-specific patch attack and temporally universal patch attack, temporal association in feature space can lead to better robustness to universal adversarial noises. 
\end{itemize}

As BEV representations begin to be deployed in real-world autonomous driving systems, we explore their robustness and ring the alarm of the safety issue in their usage. To further enhance the robustness of BEV models while preserving their accuracy in future works, temporal information could be adopted as in BEVFormer [29] and the redundant spatial features in the explicit BEV representations can be reinforced with dense supervisory signals.

\subsection*{Acknowledgments}
\footnotesize{This work was supported by the National Key Research and Development Program of China (No. 2020AAA0104304), NSFC Projects (62076147, 62071292, 61771303, 62276149, 61876001,U19B2034, U1811461, U19A2081), STCSM Project (No. 18DZ2270700), and the General Project of the Natural Science Foundation of Anhui Province (No. 1808085MF175),  Tsinghua-Alibaba Joint Research Program, Tsinghua-OPPO Joint Research Center for Future Terminal Technology. Y. Dong was also supported by the China National Postdoctoral Program for Innovative Talents and Shuimu Tsinghua Scholar Program.}

\newpage
{\small
\bibliographystyle{ieee_fullname}
\bibliography{egbib}
}

\clearpage

\appendix
\normalsize
\section{Formulations for Adversarial Attack}
\label{sec:appendix-formulations}

Adversarial attack is often formalized as an optimization problem under some certain constraints, which vary among different attack settings. Here, we give a brief introduction to the settings mentioned in the paper and formalize their objectives following the notations in~\cref{sec:setup}. 

\textbf{$\ell_\infty$ Adversarial Perturbations. }To corrupt the inputs of one frame with $\ell_\infty$ adversarial perturbations for untargeted attack, the problem is formalized as
\begin{equation}
\begin{split}
    \max_{x'} \mathcal{L}(f_\theta(x'), \hat{y}), \;
    \text{s.t. } ||x'-x||_\infty<\epsilon,
\end{split}
\label{eq:lp-attack}
\end{equation}
where $\epsilon$ is an allowed perturbation budget.

\textbf{Instance-specific Adversarial Patches. }
Formally, the instance-specific patch attack is described as
\begin{equation}
    \max_{\delta} \mathcal{L}(f_\theta((1-m)\odot x+m\odot \delta),\hat{y}),
\end{equation}
where $\delta$ is in the same space as image input $x$ and $m\in\{0,1\}^{N_c\times 1\times H\times W}$ represents binary masking matrix to appoint the location of patches with element-wise multiplication of pixels denoted by $\odot$. The masking matrix $m$ is defined according to the ground-truth 3D bounding boxes.

\textbf{Category-specific Adversarial Patches. }
The formulation of this problem is
\begin{equation}
    \max_{\delta_1,\cdots,\delta_C} \mathbb{E}_{(x,\hat{y})\sim\mathcal{D}}[\mathcal{L}(f_\theta((1-\sum^C_{j=1}m^x_j)\odot x+\sum_{j=1}^Cm^x_j\odot \delta_j),\hat{y})],
\end{equation}
where $\delta_j$ is for objects of $j$-th category in the dataset $\mathcal{D}$ which has $C$ categories in total, while $m^x_j$ denotes the binary mask for objects of the $j$-th category in the sample $x$. Similarly, the masking matrix $m_j^x$ is defined according to the ground-truth 3D bounding box coordinates.

\section{Additional Details for $\ell_\infty$ Adversarial Perturbations}
\label{sec:appendix-linf}

\subsection{Raw Data for $\ell_\infty$ Attack}
\label{sec:appendix-linf-rawdata}

The raw data for $\ell_\infty$ attack, including FGSM and PGD10, is shown in \cref{tab:appendix-fsgm} and \cref{tab:appendix-pgd}.

\begin{table*}[ht]
\centering
\small
\resizebox{1\linewidth}{!}{
\begin{tabular}{l|cccccccc}
\hline
            & Clean     & \multicolumn{1}{c}{FGSM(e=0.1)} & \multicolumn{1}{c}{FGSM(e=0.2)} & \multicolumn{1}{c}{FGSM(e=0.5)} & \multicolumn{1}{c}{FGSM(e=1)} & \multicolumn{1}{c}{FGSM(e=2)} & \multicolumn{1}{c}{FGSM(e=4)} & \multicolumn{1}{c}{FGSM(e=8)} \\ \hline
FCOS3D      & 29.8/37.7 & 25.9/34.3  & 23.4/32.5  & 19.6/29.3  & 16.9/27  & 14.9/25.4& 14.2/24.6& 13.9/24.3\\
BEVDet      & 29.2/37.2 & 17.9/26.9  & 13.7/23.1  & 8.7/17.9   & 6.1/14   & 4.7/10.7 & 3.9/10   & 3.4/9.2  \\
BEVDepth    & 33.2/40.4 & 21.6/30.4  & 16.7/27    & 10.9/21.8  & 7.9/17.7 & 6.2/14.4 & 5.3/13.7 & 4.6/9.1  \\
DETR3D      & 34.7/42.2 & 21.5/31.1  & 17/27.6    & 12.4/24.1  & 10.4/22.7& 9.7/22.1 & 10.4/22.5& 12.3/24.3\\
BEVFormer   & 37.0/47.9 & 16.2/26.9  & 12.4/23.9  & 8.2/19.9   & 6.3/17   & 5.4/16.1 & 5.6/16.3 & 6.6/17.3 \\ \hline
TransFusion & 67.2/70.9 & 64.6/69.5  & 63.3/68.9  & 61.7/68.1  & 60.9/67.6& 60.4/67.4& 60.6/67.5& 61.4/68  \\
BEVFusion   & 68.5/71.4 & 62.7/68.1  & 58.5/65.7  & 52/62.1    & 48.1/59.8& 46.2/58.8& 46.9/59.2& 50.6/61.4\\ \hline
\end{tabular}
}
\caption{\textbf{ $\ell_\infty$ FGSM attack. }mAP/NDS of FGSM attack at different $\epsilon$ settings}
\label{tab:appendix-fsgm}
\end{table*}

\begin{table*}[th]
\centering
\small
\resizebox{1\linewidth}{!}{
\begin{tabular}{l|cccccccc}
\hline
        & Clean     & PGD10(e=0.1) & PGD10(e=0.2) & PGD10(e=0.5) & PGD10(e=1) & PGD10(e=2) & PGD10(e=4) & PGD10(e=8) \\ \hline
	FCOS3D      & 29.8/37.7 & 25.4/33.6    & 21.8/30.3    & 14.6/24      & 9.3/19.1   & 5.2/13.4   & 2.6/9      & 1.2/6.4    \\
	BEVDet      & 29.2/37.2 & 17.2/26.2    & 11.6/19.9    & 5.7/11.9     & 3.3/6.8    & 1.8/5.1    & 0.8/4.3    & 0.2/0.7    \\
	BEVDepth    & 33.2/40.4 & 21/29.8      & 14.6/25.5    & 7/12.7       & 3.7/7      & 1.8/5.2    & 0.8/4.4    & 0.3/0.7    \\
	DETR3D      & 34.7/42.2 & 19.5/29.6    & 12.2/24.4    & 4.7/19.2     & 1.8/15.9   & 0.8/8.9    & 0.4/5.6    & 0.3/5.1    \\
	BEVFormer   & 37.0/47.9 & 15.7/26.6    & 10.3/22.7    & 4.1/16       & 1.6/11.7   & 0.7/8.4    & 0.3/5.3    & 0.2/3.6    \\ \hline
	TransFusion & 67.2/70.9 & 64.4/69.5    & 62.5/68.4    & 57.8/65.9    & 48.7/61.4  & 36.2/55.2  & 28.2/51.4  & 28.3/51.3  \\
	BEVFusion   & 68.5/71.4 & 62.8/68.2    & 57.3/65      & 44.1/57.7    & 30.7/50.1  & 19.1/43.3  & 11.6/38.6  & 7.5/36.1   \\ \hline
\end{tabular}
}
\caption{\textbf{ $\ell_\infty$ PGD10 attack. }mAP/NDS of PGD10 attack at different $\epsilon$ settings}
\label{tab:appendix-pgd}
\end{table*}

\subsection{Details for Transfer Attack}
\label{sec:appendix-linf-transfer}

\begin{figure}
    \begin{subfigure}[t]{0.45\textwidth}
         \centering
         \includegraphics[width=\textwidth]{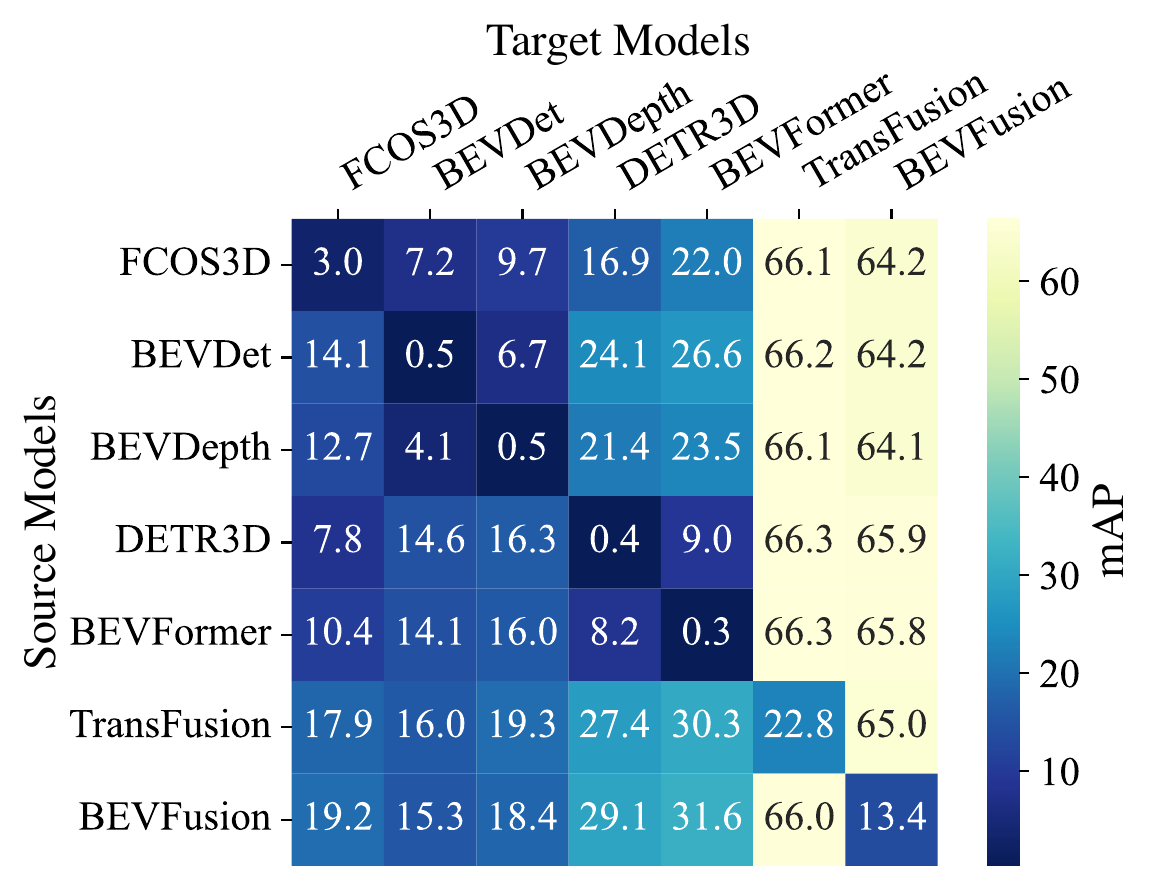}
         \caption{mAP of transfer attack}
         \label{fig:trans_mAP}
     \end{subfigure}
    \begin{subfigure}[t]{0.45\textwidth}
         \centering
         \includegraphics[width=\textwidth]{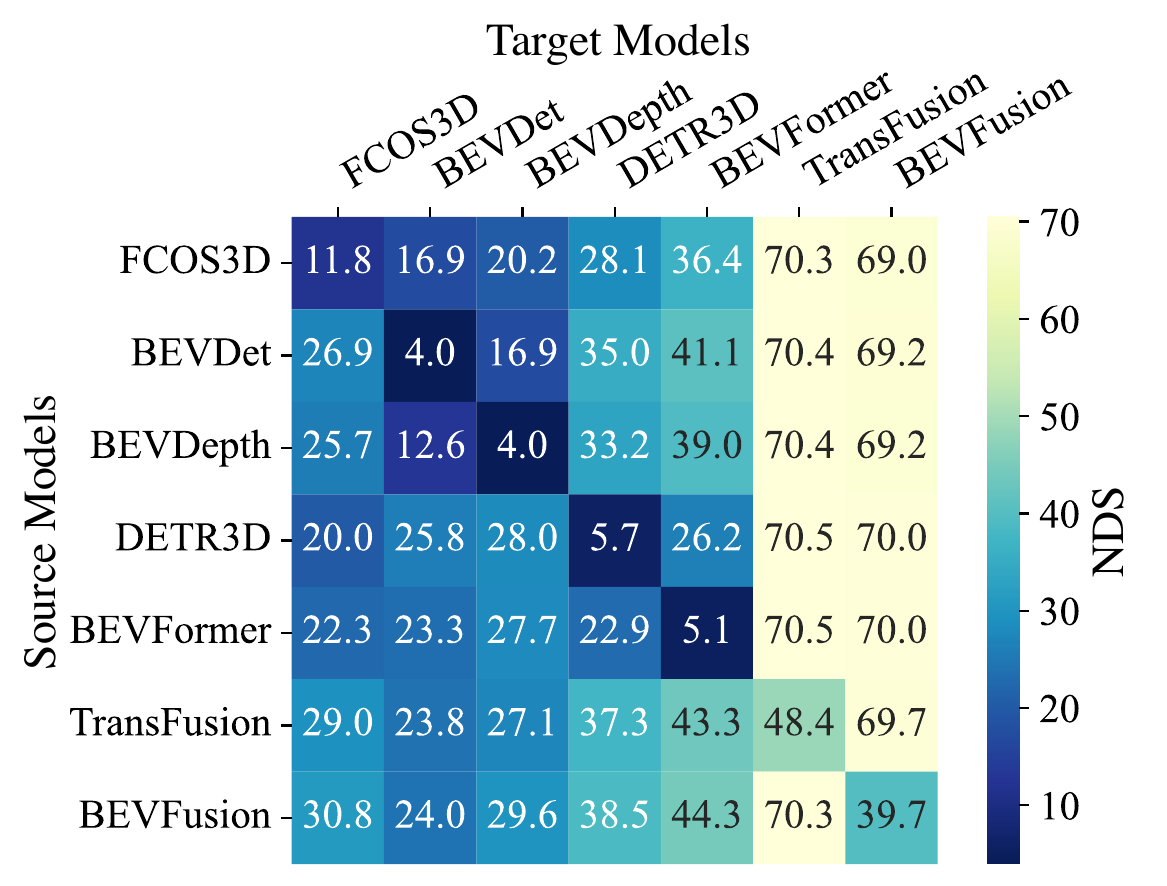}
         \caption{NDS of transfer attack}
         \label{fig:trans_NDS}
     \end{subfigure}
     \caption{\textbf{Transfer attack. }mAP, NDS of Camera-LiDAR fusion models under different adversarial perturbations added to images and point clouds.}
     \label{fig:trans}
\end{figure}

The details of transfer attack are shown in \cref{fig:trans}.

\subsection{mAP data for $\ell_\infty$ Perturbations for Fusion Models}
\label{sec:appendix-linf-fusion}
\begin{figure}
     \centering
     \includegraphics[width=0.99\linewidth]{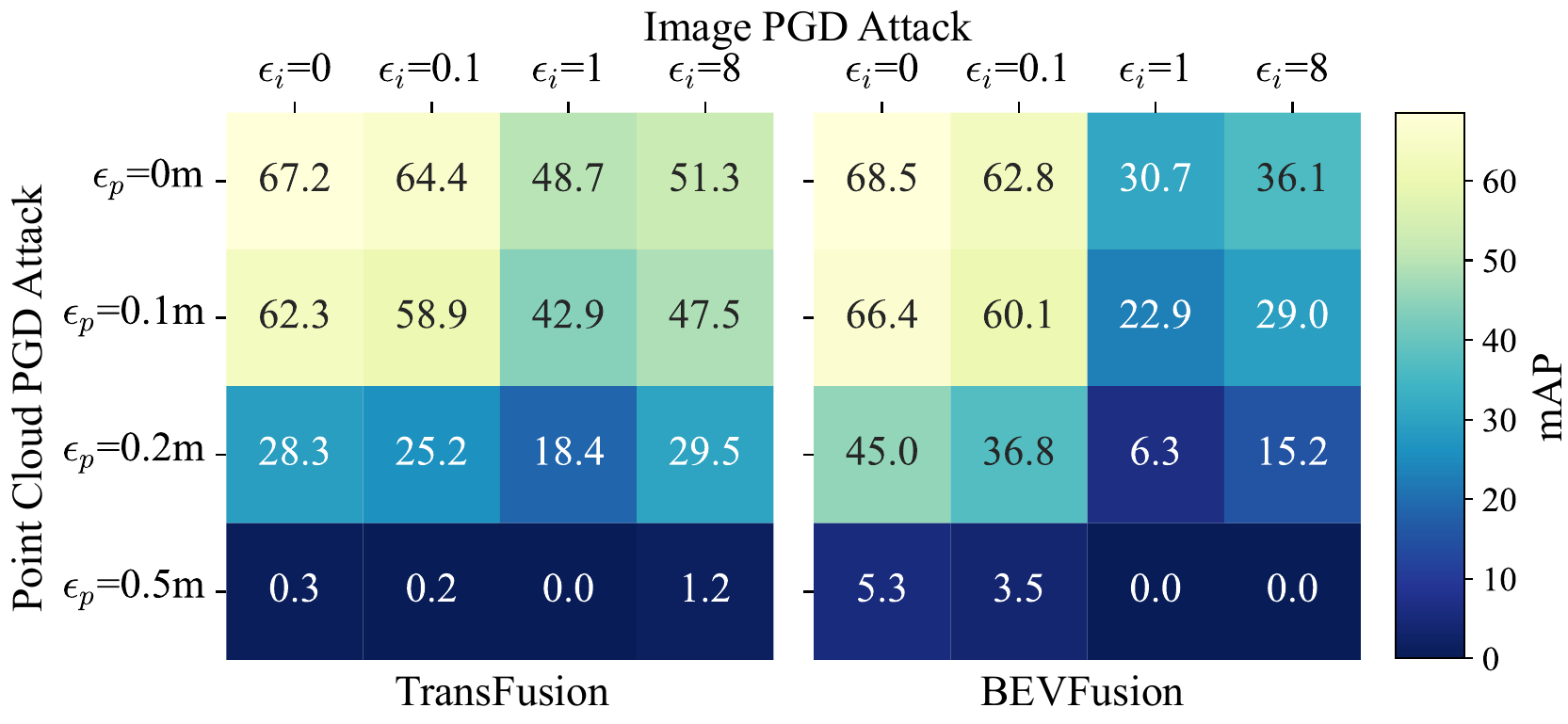}
     \caption{\textbf{$\ell_\infty$ Perturbations for Fusion Models. }mAP of Camera-LiDAR fusion models under different adversarial perturbations added to images and point clouds.}
     \label{fig:pointimg_mAP}
\end{figure}

The mAP data for $\ell_\infty$ perturbations for two fusion models, TransFusion~\cite{bai2022transfusion} and BEVFusion~\cite{liu2022bevfusion}, is shown in \cref{fig:pointimg_mAP}.

\section{mAP Data for Partial Cameras}
\label{sec:appendix-partial}
\begin{figure}
     \centering
     \includegraphics[width=0.99\linewidth]{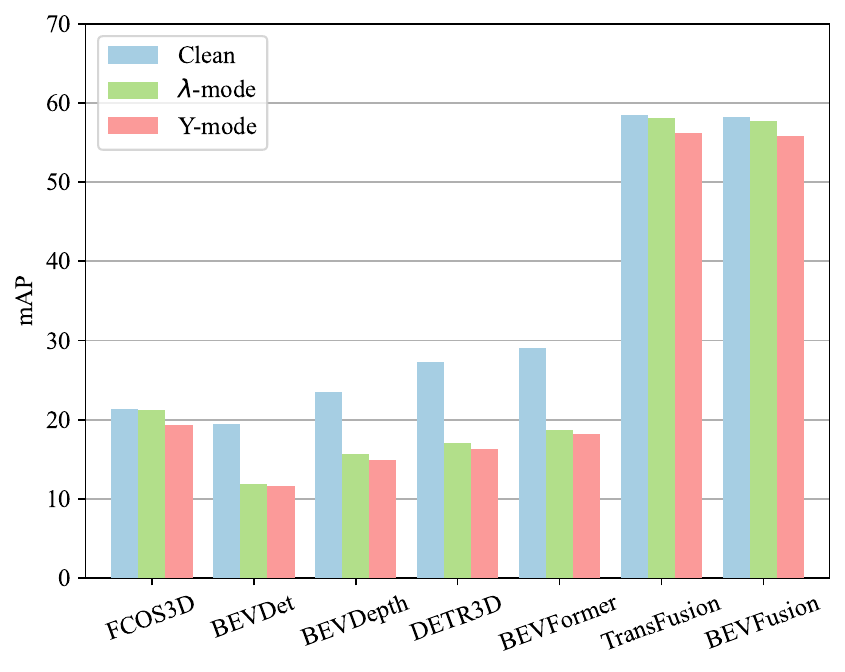}
     \caption{\textbf{Partial Cameras. }Performance with partial cameras in terms of mAP}
     \label{fig:partial_camera_mAP}
\end{figure}

The mAP data is shown in \cref{fig:partial_camera_mAP}

\section{Additional Details for 3D Consistent Patch Attack}
Full results including mAP and NDS is show in \cref{tab:overlap_1_appendix} and \cref{tab:overlap_2_appendix}.

\begin{table}[]
\centering
\small
\begin{tabular}{l|ccc}
\hline
Patch Size     & 0\% & 5\%            & 10\%      \\\hline
FCOS3D         & 21.3/32.5    & \textbf{12.9}/\textbf{22.2}        & \textbf{8.5}/\textbf{17.5}      \\
BEVDet         & 19.4/33.8    & 3.3/11.1       & 1.4/6.2         \\
BEVDepth       & 23.5/37.7    & 4.4/11.8       & 1.9/8.8        \\
DETR3D         & 27.3/39.1    & 3.5/15.2       & 1.3/10.3        \\
BEVFormer      & \textbf{29.0}/\textbf{45.0}    & 4.1/14.8       & 1.8/9.6     \\\hline
TransFusion    & \textbf{58.4}/\textbf{66.8}    & \textbf{52.6}/\textbf{63.7}   & \textbf{50.7}/\textbf{62.8}     \\
BEVFusion      & 58.3/66.4    & 36.4/54.2      & 29.5/50.8       \\\hline
\end{tabular}
\caption{\textbf{Multi-view Patch Attack. }mAP/NDS of vision-dependent models with 3D consistent patches in the cases of Multi-view Patch Attack. 0\% for clean images.}
\label{tab:overlap_1_appendix}
\end{table}

\begin{table}[]
\centering
\small
\begin{tabular}{l|ccc}
\hline
Patch Size     & 0\% & 5\%            & 10\%     \\\hline
FCOS3D         &  29.8/37.7     & 11.9/20.6      & 6.0/15.3     \\
BEVDet         &  29.2/37.2    & 3.8/12.7      & 1.7/5.9   \\
BEVDepth       & 33.2/40.4    & 6.3/17.7      & 2.5/8.7   \\
DETR3D         & 34.7/42.2    & 16.3/28.3     & 9.5/23.2    \\
BEVFormer      &  \textbf{37.0}/\textbf{47.9}  & \textbf{18.8}/\textbf{35.0}     & \textbf{11.7}/\textbf{29.0}    \\\hline
TransFusion    &  67.2/70.9       & \textbf{61.9}/\textbf{68.0}     & \textbf{58.9}/\textbf{66.4}     \\
BEVFusion      & \textbf{68.5}/\textbf{71.4}  & 54.9/63.9     & 49.1/60.7   \\\hline
\end{tabular}
\caption{\textbf{Temporally Universal Patch Attack. }mAP/NDS of vision-dependent models with 3D consistent patches in the cases of Temporally Universal Patch Attack. 0\% for clean images.}
\label{tab:overlap_2_appendix}
\end{table}

\section{Training strategies}
\label{sec:appendix_training}
We note that different training strategies could influence the model robustness to some extents and also take them into account. Though prior works on benchmarking robustness in classification\cite{dong2020benchmarking} and detection\cite{michaelis2019benchmarkingrebuttal} usually treat training strategies as part of models instead of dissociating them, we investigate the training strategies of each model including data augmentation, learning rate, optimizer, and find that there is an insignificant difference within the three comparing sub-groups of models. The detailed training schemes are summarized in~\cref{tab:model_training_stragety}.

\begin{table*}[]
\centering
\small
\begin{tabular}{l|c|c|c|c|c|c|c}
\hline
            & Optim.     & lr       & b.s. & epoch & image aug & 3D aug & GT aug  \\\hline
FCOS3D~\cite{wang2021fcos3d}            & SGD   & 2e-3 & 2*8 & 12    & -      & RandomFlip3D & - \\
BEVDet~\cite{huang2021bevdet}           & AdamW & 2e-4 & 8*8 & 24    & ResizeRotFilp      & GlobalRotScaleTrans, RandomFlip3D & - \\
BEVDepth~\cite{li2022bevdepth}          & AdamW & 2e-4 & 8*8 & 24    & ResizeRotFilp      & GlobalRotScaleTrans, RandomFlip3D & - \\
DETR3D~\cite{wang2022detr3d}            & AdamW & 2e-4 & 1*8 & 24    & PhotoMetricDistortion      & - & - \\
BEVFormer~\cite{li2022bevformer}        & AdamW & 2e-4 & 1*8 & 24    & PhotoMetricDistortion      & - & - \\\hline
TransFusion~\cite{bai2022transfusion}   & AdamW & 1e-4 & 2*8 & 20+6  & -      & GlobalRotScaleTrans, RandomFlip3D & GTPaste \\
BEVFusion~\cite{liu2022bevfusion}       & AdamW & 1e-4 & 4*8 & 20+6  & ResizeRotFilp     & GlobalRotScaleTrans, RandomFlip3D & GTPaste \\\hline
\end{tabular}
\caption{\textbf{Different training strategies.} Training strategy information of different models evaluated in this paper, including optimizer~(Optim.), learning rate~(lr), batch size~(b.s.), training epoch and different types of data augmentation strategies.}
\label{tab:model_training_stragety}
\end{table*}

\section{Different truncation levels for partial cameras}
Following KITTI~\cite{geiger2012we}, we get the truncation ratio of objects at image boundaries and examine the performance of models at different levels. Results of NDS between DETR3D and BEVFormer are shown in \cref{tab:truncation}. BEV model outperforms non-BEV model at all truncation levels for partial cameras. This is consistent with our finding in \cref{sec:partial_cam}.

\begin{table}[!h]
\centering
\resizebox{\linewidth}{!}{
\begin{tabular}{l|cccc|cccc}
\hline
\multicolumn{1}{c|}{\multirow{2}{*}{NDS}}          & \multicolumn{4}{c|}{$\lambda$-mode}           & \multicolumn{4}{c}{Y-mode}            \\ \cline{2-9}
          & All   & Easy  & Moderate & Hard  & All   & Easy  & Moderate & Hard  \\
\hline
DETR3D    & 32.0 & 37.7  & 30.0      & 29.2   & 30.6 & 35.1  & 29.0      & 27.5  \\
BEVFormer & \textbf{37.0} & \textbf{43.3}  & \textbf{39.8}      & \textbf{36.1}  & \textbf{35.5} & \textbf{40.0}     & \textbf{34.9}      & \textbf{32.3} \\
\hline
\end{tabular}
}
\caption{\footnotesize Partial results at different truncation levels for partial cameras. 
}
\label{tab:truncation}
\end{table}

\section{BEVDet/BEVDepth with high resolution}
\label{sec:appendix_resolution}
Considering that the input resolution may influence the robustness of detectors, we test this influence on BEVDet and BEVDepth. We increase the resolution from 704x256 to 1408x512 and retrain the models by ourselves due to the lack of official models.

From~\cref{tab:pgd-res} of PGD attack, we see that higher resolution improves the robustness of BEVDet and BEVDepth, but the performance is still inferior to FCOS3D. We also test them on temporally universal patch attack, and find similar trends, \eg, under 5\% patch size, the NDS of high resolution BEVDet and BEVDepth are 18.9 and 20.5, still lower than 20.6 NDS of FCOS3D. 
The conclusion is that the original resolution improves the robustness, but cannot substantially overturn our findings.

\begin{table}[]
\centering
\resizebox{\linewidth}{!}{
\begin{tabular}{l|cccccccc}
\hline
$\epsilon$             & 0      & 0.1    & 0.2    & 0.5    & 1      & 2      & 4      & 8      \\\hline
BEVDet           & 37.2 & 26.2 & 19.9 & 11.9 & 6.8  & 5.1  & 4.3 & 0.7 \\
BEVDet-HighRes   & 41.0 & 26.7 & 20.1 & 15.0 & 10.8 & 9.5  & 7.6 & 5.2 \\\hline
BEVDepth         & 40.4 & 29.8 & 25.5 & 12.7 & 7.0  & 5.2  & 4.4 & 0.7 \\
BEVDepth-HighRes & \textbf{43.9} & 31.2 & 26.4 & 18.2 & 11.0 & 9.1  & 7.0 & 4.4 \\\hline
FCOS3D           & 37.7 & \textbf{33.6} & \textbf{30.3} & \textbf{24.0} & \textbf{19.1} & \textbf{13.4} & \textbf{9.0} & \textbf{6.4} \\\hline
\end{tabular}
}
\caption{\footnotesize NDS under $\ell_\infty$ adversarial perturbations generated by PDG10.
}
\label{tab:pgd-res}
\end{table}

\section{Similar schemes when discussing temporal information}
To further investigate the effectiveness of temporal information on improving robustness, we conduct additional experiments among BEVDet \& BEVDet4D and BEVFormer-Static (re-implemented) \& BEVFormer. The results under two time-related attack settings are shown in~\cref{tab:temporal_similar_schemes}. The results confirm that temporal information fortifies robustness to universal attack along time. 

\begin{table}[]
\centering
\resizebox{\linewidth}{!}{
\begin{tabular}{cccc|cccc}
\hline
\multicolumn{4}{c|}{Category-Specific Patch~(5\%)} & \multicolumn{4}{c}{Temporally Universal Patch~(5\%)} \\ \hline
BEVD   & BEVD4D   & BEVF-S  & BEVF  & BEVD   & BEVD4D   & BEVF-S   & BEVF  \\
15.4    & 21.3      & 32.9        & 35.9      & 12.7    & 17.5      & 30.3         & 35.0       \\ \hline
\end{tabular}
}
\caption{\footnotesize Partial results (NDS) among models with temporal information.
}
\label{tab:temporal_similar_schemes}
\end{table}

\end{document}